# Persian Typographical Error Type Detection Using Deep Neural Networks on Algorithmically-Generated Misspellings

Mohammad Dehghani*, Heshaam Faili

School of Electrical and Computer Engineering, College of Engineering, University of Tehran , Tehran, Iran

Emails: dehghani.mohammad@ut.ac.ir, hfaili@ut.ac.ir

**Abstract**

Spelling correction is a remarkable challenge in the field of natural language processing. The objective of spelling correction tasks is to recognize and rectify spelling errors automatically. The development of applications that can effectually diagnose and correct Persian spelling and grammatical errors has become more important in order to improve the quality of Persian text. The Typographical Error Type Detection in Persian is a relatively understudied area. Therefore, this paper presents a compelling approach for detecting typographical errors in Persian texts. Our work includes the presentation of a publicly available dataset called FarsTypo, which comprises 3.4 million words arranged in chronological order and tagged with their corresponding part-of-speech. These words cover a wide range of topics and linguistic styles. We develop an algorithm designed to apply Persian-specific errors to a scalable portion of these words, resulting in a parallel dataset of correct and incorrect words. By leveraging FarsTypo, we establish a strong foundation and conduct a thorough comparison of various methodologies employing different architectures. Additionally, we introduce a groundbreaking Deep Sequential Neural Network that utilizes both word and character embeddings, along with bidirectional LSTM layers, for token classification aimed at detecting typographical errors across 51 distinct classes. Our approach is contrasted with highly advanced industrial systems that, unlike this study, have been developed using a diverse range of resources. The outcomes of our final method proved to be highly competitive, achieving an accuracy of 97.62%, precision of 98.83%, recall of 98.61%, and surpassing others in terms of speed.

## 1. Introduction

With the emersion of the computer era, the way we communicate has undergone a deep transformation, leading to a declined demand for traditional printed newspapers and handwritten letters. The rise of digitalization has presented humanity with manifold opportunities, but it has also introduced several challenges. As an example, in 2020, the global internet witnessed over 306 billion emails being sent worldwide (Johnson 2021), while in the United States alone, almost 2.2 trillion text messages were exchanged (CTIA 2021). Despite the fact that these statistics highlight the advantages of recent technological advancement in enhancing communication convenience and accessibility, there are certain issues that arise when natural language is employed as a communication medium. One such problem is misspelling a word, also known as making a Typographical Error (Typo). Although it seems minor, typos can have a significant and often negative impact, particularly when made by an organization or a company (Muller et al. 2019) such as customer attrition (Stiff 2012). This type of error is more prevalent in text messages and can lead to major unwanted misunderstandings (Boland and Queen 2016). In the field of Natural Language Processing (NLP), researchers have been actively working on the development of various proofreading tools to overcome this issue.

Typically, text-editing assistants are developed to focus on correcting spelling or grammar in input texts, known as spell- and grammar-checkers, respectively. The division of text-editing assistants is influenced by Kukich's categorization (1992) of orthographical mistakes into non-word, real-word, and grammatical errors. Non-word errors happen when a word is not detected in the spell-checker's lexicon (e.g., writing 'hope' as 'hoope'). Real-word errors, on the other hand, involve using a correctly spelled word from the vocabulary but in an incorrect context (e.g., writing "I lust my keys" instead of "I lost my keys"). Lastly, grammatical errors, which unlike the previous two we do not focus on in this study, occur when a sentence does not adhere to the predefined rules of a language. Spell-checkers, which primarily focus on resolving non-word errors and real-word errors, generally comprise four consecutive subtasks. First, the input text is tokenized (tokenization), meaning it is divided into individual units such as words or characters. Next, spell-checker identify potential errors within the text (error detection). Afterwards, it aims to correct previously detected errors (error correction). Finally, the spell-checker ranks and provides a list of possible candidate corrections for each misspelt word (candidate ranking). These subtasks are part of a hierarchical module, and the overall effectiveness of a spell-checker relies on the success of individual components.

Applying test-editing tasks, such as spell checking, to low-resource natural languages like Persian presents additional challenges. Persian, spoken mainly in Iran, Afghanistan, and Tajikistan, has its own unique characteristics. While its linguistic structure has remained relatively unchanged over time (Bijankhan et al. 2011), it has incorporated numerous words from Arabic (Haghdadi and Azizi 2018). Furthermore, Persian is a free word order language, meaning that the order of words in a sentence can be flexible. Additionally, Persian includes letters that have both joiner and non-joiner forms (Ghayoomi et al. 2010). These linguistic features and complexities pose challenges when developing effective text-editing tools for Persian. Indeed, Persian poses additional challenges for text-editing tasks like spell-checking. One of the challenges is the presence of various letters that are written differently but have similar sounds. This can create difficulties when modeling the language, as there is no unified set of rules for writing (Rasooli et al. 2011). Additionally, distinguishing between white spaces and pseudo-spaces (ZWNJ[1]) can be problematic (Dastgheib et al. 2016). These factors contribute to the slow pace of computational advances in Persian language processing and its inherent ambiguity (QasemiZadeh et al. 2014). Nevertheless, several studies have focused on addressing these factors and have made progress in the field of spell-checking for Persian (Dastgheib and Fakhrahmad 2019; Samani et al. 2015). In general, these studies are focused on grammar (Ehsan and Faili 2010), a specific domain (Yazdani et al. 2020), or do not offer a comprehensive approach that is highly effective. Thus, with the advancement of neural network architectures, a more accurate method can be provided and more research needs to be conducted in order to address spell-checking in Persian.

Our objective in this study is to enhance the development of more accurate Persian spell-checking systems by focusing on error detection. The effectiveness of spell-checking systems relies heavily on the accurate identification of errors, which can then be corrected. This is particularly evident in situations where Persian words contain Arabic letter, silent letters, or incorrect spacing. Once these errors are detected, they can be easily being corrected using practical techniques like predefined regular expressions patterns (e.g., correcting ‘ي’ to ‘ی’ using re.sub (r”, "", text) in Python). Hence, the detection part is crucial since the correction part can be performed quickly with only one line of code and better detection of errors results in a large number of Persian misspellings being effortlessly corrected. To achieve the goal of improving error detection and subsequently correcting Persian misspellings, this study introduces an algorithm that constructs a dataset comprising different Persian-specific typographical errors. This dataset consists of pairs of Persian words, their misspelled variants, and algorithmically assigned typographical error labels. Using this parallel dataset, the study train a Deep Sequential Neural Network to predict error types. The Deep Sequential Neural Network architecture model is designed to take input text and perform token classification, specifically classifying each word into one of 51 classes representing different types of typographical errors. By training the model on this dataset, it learns to accurately identify and classify the typographical errors associated with each word in Persian text.

This work introduces several novel aspects that distinguish it from the existing literature. One key difference is the approach to error detection. Unlike previous methods that involved manual checking for every possible Persian typographical error of each word at each time step, the developed architecture in this study utilizes a trained neural network to detect any possible error for every word by propagating forward through the network. Additionally, while dictionary-based methods have been also employed in the past to address this problem, they are not effective in detecting real-word errors and struggle to adapt to various types of misspellings, especially when encountering out-of-vocabulary (OOV) words (Dong et al. 2019). In contrast, the proposed approach in this study consider input text at both the sentence and character levels. This comprehensive approach allows the model to capture the necessary information to detect both real- and non-word errors, effectively. In previous studies, there has been a focus on developing better confusion sets to improve performance in detecting such errors. Confusion sets consists of small groups of words that are commonly confused with each other, such as “site”, “sight”, “cite”. Constructing informative and well-investigated confusion sets is crucial for the success of spell-checking systems that utilize them. Furthermore, these spell-checking systems face the challenge of selecting the correct replacement from the confusion sets in the subsequent step. Choosing the most appropriate replacement from a set of similar words requires careful consideration and can be a complex task for the system. Therefore, in this study, addressing real-word error detection and the accurate selection of replacements from confusion sets are important aspects that need to be investigated and improved to enhance the overall performance of Persian spell-checking systems. During real-word error detection, our approach intelligently executes both steps simultaneously by propagating forward our developed neural network. This makes error detection much simpler and results in a reduction in the processing time of a spell-checker. In addition to the benefits mentioned earlier, when the error type is accurately identified, it enables more effective ordering of suggestions in the candidate ranking phase of spell-checkers (candidate ranking). Furthermore, the neural architecture developed in this study is designed to detect both non-word and real-word errors simultaneously. Our comprehensive approach allows the system to handle a wider range of errors and provides a more generalized alternative compared to previous methods. By leveraging the capabilities of the neural network, the approach reduces the need for extensive hard coding and manual rule-based interventions, making the system more adaptable and flexible.

[1] Zero-Width-Non-Joiner

Overall, the contributions of this paper can be summarized as follows:

- **Introduction of FarsTypo:** The paper introduces FarsTypo, one of the largest typographical error datasets specifically designed for the Persian language. This dataset is made publicly available, enabling researchers and practitioners to utilize it for further studies and developments [2].
- **Development of a Deep Sequential Neural Network:** A novel Deep Sequential Neural Network architecture is developed to perform typographical error type detection.
- **Adaptive Algorithm for Generating Persian-Specific Errors:** The paper presents an adaptive algorithm that generates Persian-specific spelling errors.
- **A Baseline for Future Researches:** A firm baseline is established to enable future studies to compare performance.

The rest of this paper is organized as follows. In section 2, we review the existing literature on relevant spell-checking systems. Next, in section 3, we provide a detailed description of the algorithm used for error detection and introduce the dataset, FarsTypo, which is utilized for training and evaluating the system. In section 4, we describe the architecture used to perform error type detection. In section 5, we present the experimental setup conducted to evaluate the performance of the proposed approach. We also established a baseline for performance evaluation and compares the results obtained from the proposed approach with those of previous works in the field. In section 6, discussion of the research is conducted. Finally, in section 7, we conclude our work.

## 2. Related work

The emergence of digital technologies has led to a significant increase in text generation across multiple languages, styles, and formats, spanning diverse domains such as medical (Bokharaeian et al 2023) and social media (Dehghani et al. 2021). This increase in text generation has highlighted the importance of spell-checking systems, which play a crucial role in refining and improving digitally conceived content. Over the years, several spell-checking applications have been developed, each employing its own specific strategy to tackle the task. Initially, rules-based approaches (Fahda and Purwarianti 2017) and edit distance methods (Flouri et al. 2015) dominated the literature. However, more recent advancements have seen the emergence of novel approaches based on statistics (Dashti 2018; Mjaria and Keet 2018) and similarities (Kim et al., 2021). Recent studies have demonstrated the high effectiveness of machine learning and deep learning in spell-checking tasks (Ahmadzade and Malekzadeh 2021; Singh and Singh 2021). In addition, some studies have incorporated external resources to enhance spell-checking performancs. For example, WordNet (Kumar et al. 2018), Part-of-Speech (POS) tags (Damnati et al. 2018; Sakaguchi et al. 2012), Bidirectional Encoder Representations from Transformers (BERT) (Hu et al. 2020; Zhang et al. 2020), SoundEx and ShapeEx (Bhatti et al. 2015) have been utilized to perform this task.

### 2.1. Application-based

Information Retrieval (IR) applications of spell checking system have been proposed in different studies (Vilares et al. 2016; Vilares et al. 2011). The study conducted by Vilares et al. (2016) focused on investigated the effects of misspelled queries on Cross-Language IR systems. They examined the impact of automatic spelling correction and the use of n-grams in two separate studies, with a specific focus on the Spanish-to-English Cross-Language IR system.

[2] https://github.com/mohamad-dehghani/FarsTypo

The results of their evaluation revealed that character n-grams layer plays a crucial role in enhancing the robustness of the system against misspellings.

In the studies conducted by Ghosh et al. (2016) and Mei et al. (2018), the focus was on Optical Character Recognition (OCR) systems. OCR systems are used to convert printed or handwritten text into digital format. In Ghosh et al. (2016) study, the contextual information and string similarity techniques were utilized to detect OCR errors and improve retrieval performance from erroneous text. The authors evaluated their approach using the FIRE RISOT Bangla and Hindi collections, the TREC Legal IIT CDIP collection, and the TREC 5 Confusion collection.

A statistical learning model was proposed by Mei et al. (2018) for the correction of OCR errors after post-processing. To evaluate the performance of the proposed method, the historical biology book with complex error patterns was used as a test dataset. The precision, recall, and F1-score of 99.04, 98.98, and 99.01 were achieved for the proposed tokenization method.

### 2.2. Machine learning and NLP

Several studies have been conducted on the Persian language for the purpose of spell checking. Faili (2010) is one of the preliminary works that generated a collection of confusion sets for each Persian word to categorize groups of words that are likely to be confused with each other. To detect real-word spelling errors, the Levenshtein distance with a predefined threshold was used. These sets were then utilized to measure the mutual information between confusable words as a function of joint and marginal probability distributions. The precision of 80.5% and recall of 87% were obtained by the evaluation of their proposed method on data that only contained one real-word error in each sentence.

In the subsequent work by Faili et al. (2016), the initial spell-checking approach for the Persian language was further developed to incorporate the detection of grammar errors. The resulting system, called Vafa Spell-Checker (currently industrialized and known as Virastman[3]), aimed to handle errors of different types using a set of Persian grammatical error patterns. To address grammar errors, the system employed N-gram probability distribution and POS tags. Based on the F1-score reported in the study, the spell-checker, grammar-checker, and real-word error checker achieved a score of 0.908, 0.452, and 0.187, respectively.

Another spell-checker developed by Dastgheib et al. (2016) was designed to handle both real-word and non-word errors is Perspell text. The system consists of offline and online phases that incorporate semantic aspects. After normalizing and preprocessing, Perspell constructs a confusion set before ranking real-word and non-word replacement suggestions according to a unique scoring mechanism.

Furthermore, Kermani and Ghanbari (2019) proposed an efficient spell-checking system with focus on minimizing storage requirements and processing time. For this purpose, they employed Partitioning Around Medoids (PAM) for partitional clustering. The proposed spell checker has been evaluated using the Faspell spell error corpus. This corpus contains Persian misspellings serves as a benchmark for assessing the accuracy of spell-checking systems. The assessment results showed that the proposed system achieved an accuracy of %82. Additionally, the evaluation revealed that the majority of errors in the Faspell corpus were insertion errors, accounting for 42% of the total errors.

In their study, Yazdani et al. (2020) proposed an automated misspelling detection and correction system for clinical reports using n-gram language model. The evaluation of the system was conducted using radiology and ultrasound free texts. The detection rate and the correction accuracy were reported to be 90.29% and 88.56%, respectively.

In their work, Naemi et al. (2021) focused on converting informal Persian words to their formal counterparts using a spell-checking approach. They extracted two datasets from four Persian news websites, consisting of 541,296 and 1,849,927 samples, respectively. To identify the appropriate formal candidate equivalents for the informal words, the authors used statistical analysis along with correction rules. The proposed system demonstrated 94% accuracy in identifying the Persian informal words and 85% accuracy in detecting the equivalent formal words.

Persian shares many common characteristics with several other languages with low resource bases, such as Kurdish and Urdu (Sahay and Coustaty 2023). Mustafa et al. (2023) described a lemmatization (based on morphological rules) and a correction system for word-level errors using n-grams and the Jaccard Similarity for Kurdish Kurmanji Dialec. The lemmatization process achieved accuracy of 97.7% for the noun lemmatization, while verb lemmatization accuracy reached 99.3%. Additionally, the spell-checker and spell-correction achieved accuracy rates of 100% and 90.77%, respectively.

[3] https://virastman.ir/

Aziz et al. (2023) created 125,562 sentences and 2,552,735 words to study real-word errors in Urdu language. They developed a contextual spell checker using word-gram features and machine learning algorithms such as support vector machines (SVM), Naive Bayes (NB), random forests (RF), logistic regression (LR), and K-nearest neighbor (KNN). The precision, recall, and F1-score were reported as 0.84, 0.79, and 0.81, respectively. Moreover, an error correction method based on Damerau-Levenshtein distance and n-grams achieved an accuracy rate of 83.67.

### 2.3. Deep learning

Deep learning algorithms are employed in various languages spell mistake detection and correction. Lytvyn et al. (2023) developed a model for error correction in Ukrainian texts. It was found that the neural network has the ability to correct simple sentences written in Ukrainian. However, developing a complete system requires using spell checking with dictionaries and checking rules, both simple and those based on dependency parsing results or other features. In order to save computational resources, a pre-trained BERT neural network was used. Such neural networks have half the parameters of other pre-trained models and show satisfactory results in correcting grammatical and stylistic errors. Among the neural network models, the pre-trained mT5 neural network has demonstrated the best performance with a BLUE score of 0.908.

Solyman et al., (2023) has utilized data augmentation methods to enhance the proposed approach. This article suggests novel adversarial transformation approaches for data augmentation during training, which expand the distribution of valid data and utilize augmented data as auxiliary to provide new contexts. The proposed method employs a transformer. Data augmentation enhances the encoder and continuously increases its contribution by urging the grammatical error correction model to pay more attention to the encoder's textual representations during decoding. The impact of these approaches has been investigated using a transformer for grammatical error correction in Arabic as a case study. Results on the QALB-2014 dataset show an accuracy of 78%, recall of 65%, and an F1 score of 71%.

Musyafa et al., (2022) proposes an automated grammar correction model for the Indonesian language based on the Transformer architecture. Initially, a large corpus of Indonesian language is created, which can be used for evaluating subsequent grammar correction work for Indonesian. Changes are made to the CC100-Indonesian dataset, and errors are introduced using a perturbation method. Subsequently, a method based on the Transformer architecture and attention mechanism is proposed. This model has an encoder-decoder structure, with each of the four stacked blocks being identical. An average F1 score of 0.7194 and a BLEU score of 78.13 were achieved for grammatical error correction.

Hossain et al. (2021), has developed a system for checking spelling and grammar errors in Bengali language. Initially, a comprehensive and generalized monolingual Bengali corpus comprising over 100 million words is constructed using various reputable online sources. Subsequently, an extensive Bengali lexicon consisting of over 1 million unique words from this corpus is extracted. Based on this corpus and vocabulary, a combined spelling and grammar checker program is created, which simultaneously detects distinct spelling and grammar errors and provides appropriate suggestions for both. The spell checker employs the Double Metaphone algorithm to separately identify Bengali spelling errors with an accuracy of 97.21%. For grammar errors, errors are detected based on the language model probability, namely, bigram and trigram combinations, and generates suggestions based on cosine similarity criterion with an accuracy of 94.29% separately.

In Lin et al. (2021) study, a grammatical error correction framework for the Indonesian language has been introduced. This framework approaches grammatical error correction as a classification task. The method integrates various language embedding models and deep learning models to correct 10 types of POS errors in Indonesian text. Additionally, an Indonesian dataset has been created, which can be used as an evaluation dataset for Indonesian grammatical error correction research. The results showed that the Long Short-Term Memory (LSTM) word embedding-based model achieved the best performance with an F1 score of 0.551.

For Persian language, Golizadeh et al. (2022) proposed the Seq-GAN framework, which aims to generate Persian sentences containing common grammatical and semantic errors. The purpose of this framework was to train Persian Grammatical error detection systems. The authors used a dataset consisting of 8000 samples to train the generator According to the reported results, the Seq-GAN framework achieved a BLEU-2 score of 64.5%, a BLEU-3 score of 44.2%, and a BLEU-4 score of 21.4%.

In the study by Irani et al. (2022), a method for correcting bilingual Arabic and Persian spelling errors was proposed. The approach utilized a conditional random field recurrent neural network (RNN). The researchers had access to a dataset consisting of 220,000 sentences containing both Arabic and Persian contents. In another study, Elahimanesh et al. (2022) utilized this dataset to develop a method based on n-gram language model. They manually added a variety of spelling errors to the data. A precision, recall, and F1-score of 92.44%, 91.62%, and 92.03% were achieved using test data, respectively.

### 2.4. Persian datasets

Several datasets for Persian spell-checking have been proposed. Similar to Aspell[4], FAspell[5] (QasemiZadeh et al. 2006) is one of the best known Persian spell-checking datasets which contains 5,050 word-pairs spelt correctly and incorrectly by school students or professional typists. Additionally, another source has been used comprising about 800 pairs extracted from an OCR system. The Persian Spell-checker[6] is another instance that contains 32,325 verbs and 3,172 adjective pairs extracted from the Moin Encyclopedia and the Persian Wikipedia[7]. Most recently, Oji et al. (2021) introduced PerSpellData, which containes 3.8 and 2.5 million pairs of non-word and real-word error-containing sentences of both formal and informal text. It comprises a wider range of topic diversity than previously available datasets. This public dataset is less synthetic, where most errors come from actual mistakes made by users on the Virastman platform (extracted from system logs).

Although there are a number of Persian error datasets available, they do not consider all kinds of errors. Our objective was to develop a system that is capable of identifying any type of error (common or unusual). Therefore, we generate a dataset with a broad range of errors, which is an advantage of our method.

## 3. FarsTypo

In this section, we will explain the process of developing the FarsTypo dataset and highlight its distinguishing features that set it apart from existing spell-checking datasets. Our primary objective is to create a parallel dataset consisting of Persian words in both their correct and incorrect forms for the purpose of error type detection. However, accomplishing this goal presents us with numerous challenges:

- Given the importance of POS tags in spell-correction systems as emphasized by previous studies Mizumoto and Nagata (2017) and Sakaguchi et al. (2012), it is crucial to incorporate them in our approach. Many existing spell-correction systems have heavily relied on these tags, making it essential to consider them as a valuable resource for future research in the field of spell-correction.
- Dealing with non-word errors and real-word errors requires different processing scales, specifically word-level and sentence-level approaches, respectively.
- The diversity of a dataset is strongly correlated with the generalization success of systems that utilize it.

To tackle the mentioned challenges, our FarsTypo dataset employs the following strategies.

- To address the first challenge, we use four different datasets named Persian Discourse Treebank (PerDTB) (Mirzaei and Safari 2018), Uppsala Persian Corpus (UPC) (Seraji 2015), ParsNER-Social (Asgari-Bidhendi et al. 2021), and ParsNER-Wiki[8]. These datasets contain words annotated with their respective POS tags, ensuring professional-level annotation.
- Regarding second challenge, all of these datasets share a key characteristic, which allows us to gain insight into both the word- and sentence-level information. These datasets consist of distinct sentences, where each sentence maintains the correct order of word presentation within the sentence.
- Moreover, these datasets also provide a solution to the third challenge since they cover a wide range of genres, such as news, law, medicine, culture, art, sports, gaming, economics, and fiction in both informal and formal linguistic styles.

[4] http://aspell.net/
[5] http://pars.ie/lr/faspell_dataset
[6] https://github.com/reza1615/Persian-Spell-checker
[7] https://fa.wikipedia.org
[8] https://github.com/majidasgari/ParsNER/tree/master/persian

Table 1 provides an overview of the dataset indicating that it consists of 3,450,918 words with chronologically POS-tagged. These words are distributed across exactly 117,000 sentence. Within the dataset, there are 101,747 unique words and 62 distinct POS tags. Fig. 1 provides a visualization of the word length statistics within our dataset. It reveals that words containing two characters are the most common, constituting 18.8% of the dataset (647,858 non unique words). Following that, words with four characters make up 16.7% of the dataset (575,723 non-unique words). The frequency of words with one, three, and five characters is nearly identical, accounting for approximately 14% of the dataset. Fig. 2 provides an overview of the distribution of POS tags within dataset. It indicates that the largest category of tags in the dataset is nouns, accounting for 40.01% (1,382,905) of the dataset. Following that, pronouns are the second most common tag, representing 10.5% of the dataset. Verbs and adjectives are also prominent, with frequencies of 9.8% and 9.5%, respectively.

**Table 1** Statics of FarsTypo dataset.

| Item | Count |
| --- | --- |
| Words | 3,450,918 |
| Sentences | 117,000 |
| Unique words | 101,747 |
| Distinct POS tags | 62 |

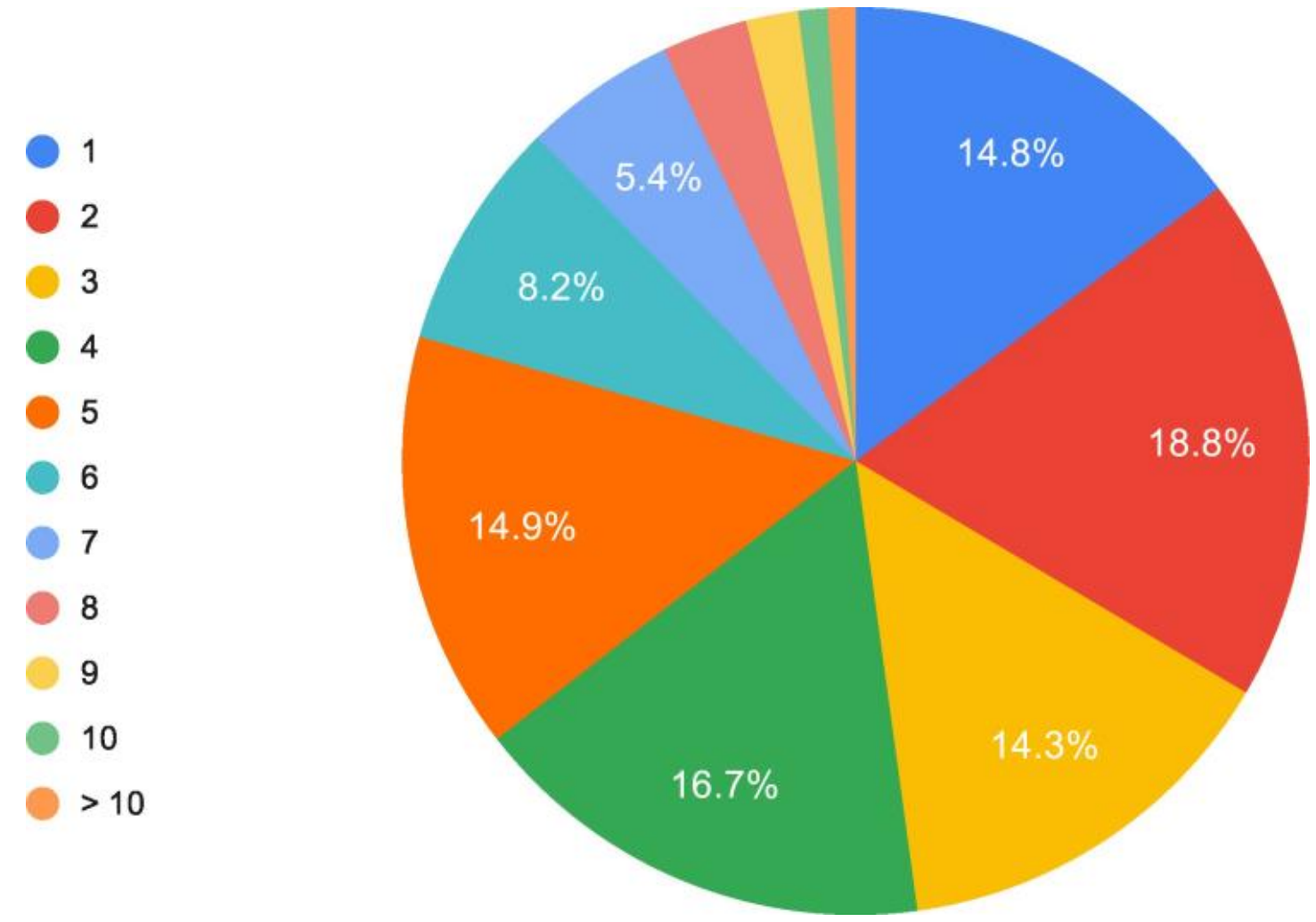

**Fig. 1** The distribution of words in the FarsTypo.

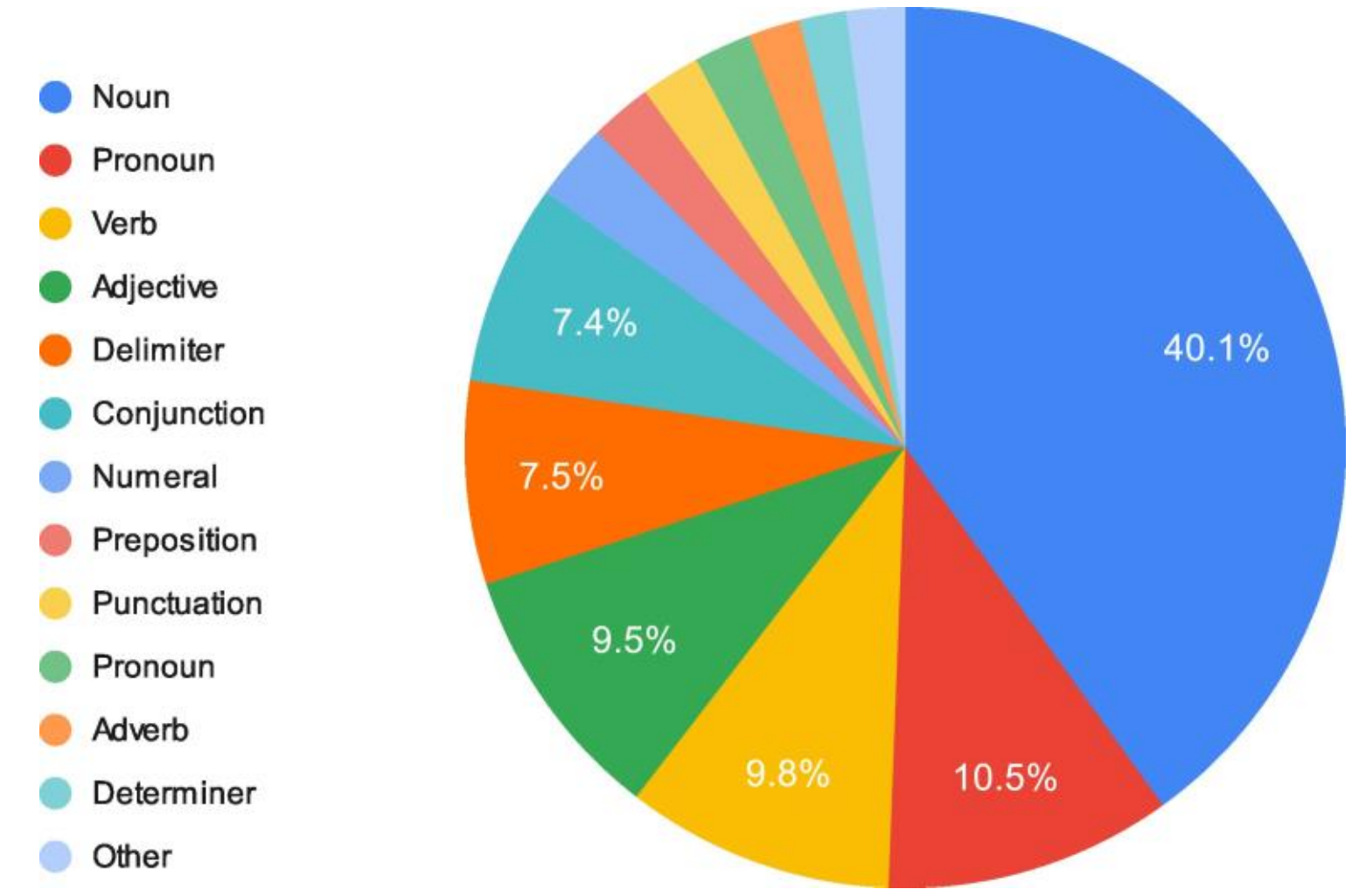

**Fig. 2** The distribution of POS tags in the FarsTypo dataset.

In the subsequent steps, we will introduce various types of synthetic typographical errors to each word in the dataset. Unlike previous studies that mainly focused on random character insertions, deletion, substitution, or transposition, this study also explores common types of mistakes made by Persian writers. It was our intention to develop a comprehensive system that would be able to detect a wide range of errors, so we took into consideration both common and uncommon errors. Subsequent sections will provide detailed descriptions of each error type.

(1) **Insertion:** This error happens when a character is accidentally inserted into a word (عالی /aali/ (excellent) ⇒ عالیس). In real-world scenarios, this error often occurs when the author inadvertently presses an adjacent button along with intended one on the standard Persian keyboard (Kashefi et al. 2013). To simulate this error, we consider the adjacency of characters on the keyboard, taking into account all sides of each button. This approach prevents randomness from having a major impact and generates more natural errors. It is important to note that when inserting characters randomly, we avoid placing identical characters next to each other, as this would result in a repetition error, which will be discussed separately.

(2) **Deletion:** This error occurs when a character is accidentally skipped or omitted in writing (دار /daar/ (scaffold) ⇒ در).

(3) **Substitution:** When the wrong keyboard button is pressed instead of the intenden one, a substitution error happens ( خال/khal/ (mole) ⇒خار ). In this error type, different character is substituted for the correct one. To simulate this error, we also consider the adjacency of buttons on the keyboard, which helps in reducing randomness and generating more realistic errors. Substitution often arise from human errors during the writing process.

(4) **Transposition:** Another potential error, especially when writing quickly, is to write two characters back to back (روز/rooz/ (day) ⇒زور ).

(5) **Similarity Confusion:** Typographically, this error type can be further divided into sound similarities and shape similarities.

   (i) Sound Similarities: In this case, two letters have the same sound, but are written differently (منظومه/manzoumeh/ (system) ⇒ منضومه). In Persian, there are several letters that fall into this category. As an example, the letter 'z' is written in four different forms: 'ز, ذ, ظ, ض', despite having the same sound. It is important to note that not all languages exhibit this phenomenon, and it is particularly relevant in Persian. English, for example, does not possess such variations.

   (ii) Shape Similarities: This error occurs when a word is misspelled due to the appearance of certain letters ( ثانویه/saanaviy[y]e/ (secondary) ⇒ تانویه).

(6) **Repetition:** Due to finger vibrations or hesitation, a button can be either pressed several times or held down while typing, resulting in repetitive typing of the same character (داودی/davodi/ (Daudi) ⇒ داوودی). This error type is distinct from the insertion error mentioned earlier, where an adjacent button is mistakenly pressed.

(7) **Spacing:** In Persian, words are typically separated by white spaces, while prefixes and postfixes are separated by pseudo-spaces. However, since letters within words are written continuously in Persian, it is possible for errors to occur in the spacing between these letters. There are six potential spacing mistakes that can be made, as each of these spaces can be confused with another:

- (i) Changing a pseudo-space to a white space ( بی‌حوصله/bihõsele/ (impatient) ⇒بی حوصله ).
- (ii) Changing a pseudo-space to an empty space ( بی‌حوصله/bihõsele/ (impatient) ⇒بیحوصله ).
- (iii) Changing a white space to a pseudo-space ( کتاب خواندن /ketaab khaandan/ (to read a book) ⇒ کتاب‌خواندن ).
- (iv) Changing a white space to an empty space ( کتاب خواندن /ketaab khaandan/ (to read a book) ⇒ کتابخواندن ).
- (v) Changing an empty space to a pseudo-space ( داستان /daastaan/ (story) ⇒ داست‌ان ).
- (vi) Changing an empty space to a white space ( داستان /daastaan/ (story) ⇒ داست ان ).

(8) **Persian to Arabic Conversions:** There are certain characters in Persian that are unique to the language and are written differently compared to Arabic. It is not uncommon for Arabic characters to be used instead of Persian characters when encoding. Despite the fact that some characters are almost identical in shape, they are written slightly differently in Persian. Therefore, it is crucial to be attentive to the specific Persian writing style. A reliable spell-checking system should be capable of identifying these characters accurately. While the letters of Persian and Arabic are mostly the same, there are a few differences. This can lead users to use an Arabic keyboard instead of Persian one, especially if the latter is not available. We will focus on the three most common scenarios where the two keyboards may be mixed:

- (i) Writing ‘ک ‘ as ‘ك ‘ ( بی‌باک /bibaak/ (audacious) ⇒ بی‌باك ).
- (ii) Writing ‘ی ‘ as ‘ي‘ ( برای /baraaye/ (for) ⇒ براي ).
- (iii) Writing ‘ه ‘ as ‘ة‘ ( پوشیده /pooshide/ (covered) ⇒ پوشیدة ).

(9) **Silent Letters:** Another common mistake made by Persian writers is the omission of silent letters in certain words. For example, ‘و’ /v/ is silenced in a word like خواننده/khaanande/ (singer), leading the writer to mistakenly writing it as خاننده. This error type can also be seen as having a similar effect to that of a deletion error. However, due to the linguistic and cognitive factors involved in this particular case, we consider this prevalent error to be distinct enough to be categorized as another typographical error. To simulate this error, the word is first checked to determine if it shares the same lemma as some of the most commonly used Persian lemmas that contain silent characters. If a match is found, the silent character is eliminated from the word.

(10) **Other prevalent errors:** There are also several other errors that Persian writers make. For instance:

- (i) One common mistake Persian writers often make is the confusion between two frequently used verbs, گذاردن /gozaardan/ (to put/allow) and گزاردن /gozaardan/ meaning (to perform). These verbs have similar pronunciations but convey different meanings. The main reason for this mistake is the presence of the letter /z/ in both verbs. The similarity in pronunciation often leads native speakers to make errors in writing. For example, the word قانون‌گذار /ghanoongozar/ (legislator) may be incorrectly written as قانون‌گزار.
- (ii) The Persian question mark ’ ؟’ is frequently mistaken with its English counterpart ’ ? ’ .
- (iii) Due to their similarity, the Persian comma sign ’ ،’ is sometimes written in English as ’ , ’.
- (iv) Due to the prevalence of Arabic letters in Persian words, ’ ئ’ (Hamze) and ’ ی ’ /y/ are occasionally confused by writers ( ویدئو /vide’o/ (video) ⇒ ویدیو, جزئیات / jozeiyat/ (details) ⇒ جزییات ).

Whereas these errors are categorized into ten general groups, we break them down into detailed error classes. According to our analysis of a large number of datasets and prior studies, it is observed that Persian writers typically make utmost two typos per word. This finding suggests that when more errors occur, the writer is more likely to notice them and correct them later. As a result, in simulating typographical errors, we applied either one or a combination of two errors to a given word. In the scenario where two errors are applied to a word, it is worth noting that both errors can be of the same type, such as two insertions or two deletions. This results in a total of 50 classes representing various combinations of errors, along with an additional class called N/A for words where no errors have been applied. There could have been more classes designed, however, certain combinations of two error types cannot be applied correspondingly to a word. Therefore, it is necessary to carefully consider the validity and meaningfulness of the error combinations to ensure they accurately reflect the types of mistakes commonly made by Persian writers.

As inputs to algorithm 1, we created a list of detailed modulated errors ($Y$), a list of words in our dataset ($D$), a floating-point number ($s$) determining the percentage of $D$ that we intend to apply errors to it, and an integer ($m$) representing the maximum number of errors allowed to apply on each word. According to the algorithm, a random subset of available candidate words ($C$) is selected from $D$ on which errors will be applied in further steps. As the algorithm traverses the list of words $D$, if a word is also a member of the candidate subset $C$, a random number $j$ is generated.

This random number $j$ represents the number of attempts that the algorithm will make to apply an error to that word. The range of the random number $j$ is from one up to the character length of the word. This means that $j$ can take values from 1 to the number of characters in the word, inclusive. As the inner loop starts, the algorithm intends to randomly pick an error module for $j$ number of times. For each successful attempt to apply an error to a word, the counter $i$ is increased by one. This counter is initially set to zero before the inner loop begins. The rationale behind categorizing attempts into successful and unsuccessful is that not every error can be applied to a certain word. In برادر /baraadar/ (brother), for instance, no silent letter is found to be changed if the algorithm so chooses. This counter is further used as a condition to break the inner loop when the maximum number of errors allowed per word is reached. Lastly, the misspelt word is added to the output list $T$ alongside a list of all its successful error modules ($Y_n$). In cases where the word is not a member of $C$, the word will directly be appended to the output list and would be tagged as N/A. After set $s$ and $m$ to 0.25 and 2, a total of 2,599,178 words remained unchanged, whereas 851,740 misspelt words were generated among which 672,022 and 179,718 words contained one and two errors, respectively (Table 2 and 3).

**Table 2** Generated misspelt words by proposed algorithm.

| **Item** | **Count** |
|---|---|
| Unchanged words | 2,599,178 |
| Generated misspelt words | 851,740 |

**Table 3** Word's error number.

| | **One** | **Two** |
|---|---|---|
| **Number of errors in words** | 672,022 | 179,718 |

Indeed, while the algorithm aims to mimic the randomness of typographical errors, it is important to consider the potential negative consequences of this uncertainty. This highlights the significance of post-processing where randomly generated misspellings are qualitatively controlled. More specifically, we check for empty, recurrent, and identical outputs.

- If the randomly generated number $j$ happens to be equal to a word's character length, and the selected error module for that iteration is deletion, it could indeed result in an empty output. This situation can occur, although the probability of it happening in each iteration is relatively low.
- Regarding the second, it is probable that the randomly selected error modules can cancel out each other's effects. For instance, in the case of خواهر/khaahar/ (sister), the application of the deletion module followed by the addition of the same character can indeed result in the output being indistinguishable from the original word. In this scenario, if the deletion module first removes the starting character, resulting in "واهر," and then the addition module randomly selects to add "خ" (kh) to the beginning of the misspelt word, the final output would be "خواهر," which is identical to the original word.
- Finally, some misspellings could be generated with identical spellings but different tags, which can complicate the training and testing of deep learning models in the further stages.

During post-processing, these phenomena are filtered (5,810 samples), leaving only valid misspellings in the $T$ set.

---

**Algorithm 1:** Generating Typographical Errors

---

**Data:** $D, Y\ , s, m$
**Result:** $T$
$T$ = [ ];
$p = s \times size\ (D)$;

$C$ = select $p$ random words from $D$; // Select a portion of the given dataset;
**for each** $w \in D$ **do**
**if** $w \in C$ **then**
$l$ = $length$ ($w$);
$j$ = $random$ (1, $l$ + 1); // Select the number of misspellings to apply to $w$;
$i$ = 0; // Initialize a counter to store the number of misspellings made;
$w_n$ = $w$; // Store a copy of the original word to further apply typographical errors;
$Y_n$ = [ ]; // Initialize a list of error types that will be successfully applied to $w$;
**for each** _ $\in$ $range(j)$ **do**
$r$ = $random$ (1, $length$ ($Y$ ));
$w_n$ = $Y$ [$r$] ($w_n$); // Select a random typographical error as $Y$ [$r$] and apply it to $w$;
**if** $w \neq w_n$ **then**
// Check if the error was successfully applied to $w$;
$i$ + +; // Increment the number of successfully-applied misspellings;
$Y_n$.$append$ ($Y$ [$r$]); // Add typographical error $Y$ [$r$] to the list of successfully-applied errors;
**if** $m$ == $i$ **then**
// Check if the number of changes we applied to word $w$ has reached the maximum number of errors allowed per word.;
$break$;
**end**
**end**
**end**
$T$.$append$ ([$w_n$, $Y_n$]);
**end**
**else**
$T$.$append$ ([$w$, N/A]);
**end**
**end**

A sample of our dataset is shown in Table 4 where three samples of the FarsTypo dataset are indicated, each representing a sentence taken from baseline datasets that have been previously discussed. Each sentence in our dataset contains an ID that distinguishes it from the rest. Including the baseline dataset information in the Source column enhances the transparency, traceability, and reproducibility of the research or application using the generated misspellings. The number of rows in each sample, indicating the number of chronologically-ordered words with their respective POS tags, provides valuable context for understanding the structure and linguistic features of the dataset. Using our algorithm, the misspellings and error types are generated and written respectively in the Misspelt Word and Typo Type columns, parallel to the original words. These samples clearly display a diversified dataset that covers a broad spectrum of linguistic styles and topics is valuable for creating robust and adaptable language models and NLP systems.

**Table 4** Different samples of the FarsTypo dataset. (wiki = ParsNER-Wiki, social = ParsNER-Social, PDTB = PerDTB, UPC = Uppsala Persian Corpus)

| #Row | ID | Word | POS Tag | Source | Misspelt Word | Typo Type |
|---|---|---|---|---|---|---|
| 1 | 8 | بیشتر | DET | wiki | بیشتتر | repetition |
| 2 | 8 | جمعیت | N | wiki | جمع یت | white space |
| 3 | 8 | این | DET | wiki | ال‌ین | insertion, pseudo space |
| 4 | 8 | ولایت | N | wiki | ولا ی ت | white space, white space |
| 5 | 8 | را | P | wiki | ژع | shape similarity, substitution |
| 6 | 8 | تاجیک‌ها | N | wiki | ت‌اجیک‌ها | pseudo space |
| 7 | 8 | تشکیل | N | wiki | تکیل | deletion |
| 8 | 8 | می‌دهند | V | wiki | می‌ذهند | shape similarity |
| 9 | 8 | . | PUNC | wiki | . | N/A |
| 1 | 35326 | گروه | Ne | social | گـروه | pseudo space |
| 2 | 35326 | نرم | AJe | social | نرم | N/A |
| 3 | 35326 | افزاری | N | social | اافففزاری | repetition, repetition |
| 4 | 35326 | دانشکده | Ne | social | دانشکد | deletion |
| 5 | 35326 | صنایع | N | social | صنای ع | white space |
| 6 | 35326 | برگزار | AJ | social | بررگززار | repetition, repetition |
| 7 | 35326 | می‌کند | V | social | می‌کند | N/A |
| 8 | 35326 | : | PUNC | social | : | N/A |
| 1 | 35305 | نه | ADV | PDTB | نه | N/A |
| 2 | 35305 | خودم | PR | PDTB | خودم | N/A |
| 3 | 35305 | را | POSTP | PDTB | ا | deletion |
| 4 | 35305 | می‌بینم | V | PDTB | می‌بینم | N/A |
| 5 | 35305 | و | CONJ | PDTB | و | N/A |
| 6 | 35305 | نه | ADV | PDTB | نه | N/A |
| 7 | 35305 | به | PREP | PDTB | تهخ | insertion, shape similarity |
| 8 | 35305 | راش‌ها | N | PDTB | راش‌ها | N/A |
| 9 | 35305 | نگاه | N | PDTB | نکاه | substitution, shape similarity |
| 10 | 35305 | می‌کنم | V | PDTB | می‌کنم | N/A |
| 11 | 35305 | . | PUNC | PDTB | . | N/A |

## 4. Proposed Method: A Deep Sequential Neural Network

Our approach to predicting error types involves training a neural network on a parallel dataset (denoted as $T$) consisting of misspelt words and their corresponding typographical error types. In this dataset, there are a total of 51 possible combinations of tags, each representing a separate class of error. To handle the prediction of error types, we frame it as a token classification task, where each word can have a specific combination of errors leading to misspelling. As mentioned previously, each error is possible to be applied to a word more than once, which prevents us from performing a multi-label classification. Moreover, while using a neural classifier can perform well in predicting non-word errors, it poses challenges in identifying real-word errors. Addressing this challenge requires careful consideration.

It should be noted that preprocessing steps, such as stop word removal (Dehghani and Manthouri 2021), are not performed on the created dataset. This is because errors in stop words might occur and need to be identified. The chronological order of our baseline datasets plays a crucial role in here. It allows us to develop a Deep LSTM Neural Network and train it on sentences where words are arranged in sequential order. These sentences contain words that may or may not have typographical errors and may or may not be correctly placed within the sentence. By training the LSTM network on such structured data, the hidden layers of the neural network can learn the effects of both real-word and non-word errors in sentences. This enables the network to capture the patterns and relationships between words and errors, enhancing its ability to predict error types effectively.

Our proposed architecture, as depicted in Fig. 3, has been developed using the Keras library (Chollet, 2015). This architecture comprises multiple layers, including embeddings (word and character), spatial dropout, bidirectional LSTM, and a dense layer.

***Word embedding layer***

In the proposed architecture, the representation of inputs is a crucial aspect of developing an effective neural network. To address this, the first layer of the architecture consists of embeddings. Word embeddings are essentially dense vector representations of words (Mao et al. 2016). They are an arrangement of real-valued numbers representing the semantic and syntactic information of words and their context, in a format that computers can understand (Naseem et al. 2020). The representation is a real-valued vector that encodes the meaning of the word in such a way that words that are closer in the vector space are expected to be similar in meaning (Patil et al. 2023; Zaland et al. 2023). The Embedding layer is used to represent the input data in a dense vector space. The input data is a sequence of integers, and each integer is mapped to a fixed-size vector. Almost all modern NLP applications start with an embedding layer.Word embeddings have been used in various NLP applications, such as text classification (Dogra et al. 2022), sentiment analysis (Shukla et al. 2023), named entity recognition (Srivastava et al. 2023), and topic modeling (Kinariwala et al. 2023).

Word embeddings are commonly used to map correct words to points in an n-dimensional space, where similar words are close to each other. However, since the primary focus of the study is on OOV words. representing a significant number of them using word embeddings may lead to model degradation (Y. Kim, Jernite, Sontag, & Rush, 2016). To overcome this limitation, character embeddings are utilized, as they have shown superiority in representing OOV words, aligning with the objectives of the study. Given that the task involves both correct and incorrect words, a combination of word and character embeddings is leveraged. The word and character embeddings are separately processed through the Embedding and TimeDistributed Embedding layers, respectively. These processed embeddings are then concatenated before being passed into the next layers.

***Spatial Dropout layer***

Spatial dropout is a type of dropout regularization technique that is used in convolutional neural networks (CNNs) to prevent overfitting (Tompson et al. 2015). Unlike traditional dropout, which randomly drops out individual neurons, spatial dropout drops out entire feature maps (Lee et al. 2020). This is because feature maps in CNNs are highly correlated, and dropping out individual neurons may not be effective in preventing overfitting. By dropping out entire feature maps, spatial dropout can help prevent co-adaptation of feature maps and improve the generalization performance of CNNs (Park et al. 2017; Lee et al. 2020). Spatial dropout can also be applied to LSTM based neural networks (Choudakkanavar and Mangai 2022). In our architecture, to address the risk of overfitting, Spatial Dropout layers are included in the architecture before bidirectional LSTM. These layers randomly drop out entire 1D feature maps, reducing interdependence between neighboring features and improving the model's ability to generalize.

***Bidirectional LSTM layer***

The LSTM network is a type of RNN designed to overcome the problem of vanishing gradients (Abandah et al. 2021). LSTM cells have a more advanced internal architecture compared to basic RNN cells due to a set of gates that govern the operation of each individual cell. The functioning of LSTMs is based on the concept of cell state gates, which control the removal or addition of information (Alom et al. 2019). The Bidirectional LSTM is a form of LSTM that exploits both future and past contexts (Abandah et al. 2021). The architecture of a bidirectional LSTM comprises two unidirectional LSTMs that process the sequence in both forward and backward directions. The output of each LSTM is combined to form the final output of the bidirectional LSTM (Huan et al. 2022). This architecture enables the model to leverage information from both directions and make more accurate predictions. Bidirectional LSTM is the next layer and used to train the model.

***Dense layer***

Lastly, a TimeDistributed Dense layer is added to the network for token classification of error types. The TimeDistributed layer in keras ensures the same operations are applied to each temporal slice of the input. When the TimeDistributed layer is applied to the input, the same function will be applied at every step of the input. Consequently, after division of the input into windows, each window is subjected to the same embeddings, special dropout, Bidirectional LSTM, and dense operations (Santacroce et al. 2020). The rationale behind using a TimeDistributed output layer is to perform classification in each time step, enabling us to predict the error types of many words in an input sentence. In our mode, TimeDistributed Dense layer applies a dense layer to each element in the sequence to produce class probabilities. The Softmax activation function is applied to each word, enabling the model to predict the error type associated with multiple words in an input sentence.

It is worth mentioning that N/A can also be predicted for those words containing no errors so that a correctly written word can also be identified by the model.

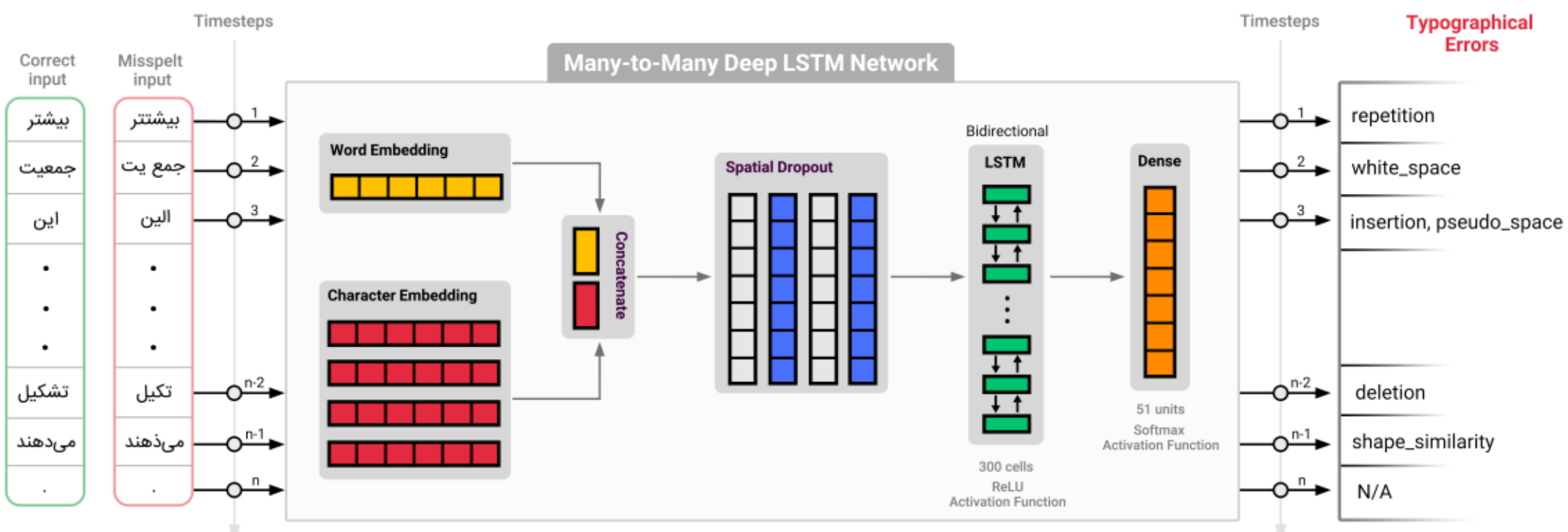

**Fig. 3** The architecture of final model.

## 5. Experiments & Results

Using different layers and text representations, we will explore various neural architectures to establish a baseline for the task of Persian spelling error type detection. The dataset used in the study contains 3,450,918 parallel words and error types, which are divided into 60%, 20%, and 20% for training, validation, and testing. Table 5 lists the parameters used to train the model. The maximum length of a sequence is set at 30 words, and the model is trained in 30 epochs with a batch size of 256. The Adam optimizer is used to adjust the network's parameters and minimize the loss function, which is calculated using categorical cross entropy. The learning rate parameter has been set to 2e-5.

**Table 5** Model training parameters.

| **Parameter** | **Value** |
|---|---|
| Maximum sequence length | 30 |
| Epochs | 30 |
| Batch size | 256 |
| Optimizer | Adam |
| Loss | Categorical cross entropy |
| Learning rate | 2e-5 |

Our goal is to develop a better final combination by comparing different architectures across different components. We experiment with both word and character embeddings, either by learning one through the model or by leveraging a pre-trained one. Table 6 and 7 respectively demonstrate the efficiencies of word-level and character-level models. The results show that, in general, word-level models have outperformed character-level models, indicating the significance of n-dimensional word representations in identifying errors. However, as discussed previously, such representation is prone to failure in non-word scenarios where a character embedding aids the model to better digest non-word errors. In addition, character embeddings performed much faster during training and testing. Although no pre-trained character embedding weights are available in Persian, fastText (Bojanowski et al. 2017) is mainly used as a pre-trained word embedding, which provides researchers with Persian word vectors. However, this model was trained on normalized Persian Wikipedia text, which lacks diversity in both genre and linguistic style. As shown in Table 6, the difference between a model leveraging fastText pre-trained word vectors and a model with trainable randomly initialized weights is minor.

Another factor to be compared between Tables 6 and 7 is how different layers have contributed either positively or actively towards the obtained results. The Vanilla RNN performed fastest in both training and testing, making the model light and more feasible to work with. However, this comes at the cost of an inefficient model. In contrast, the bidirectional layers served relatively more accurately using either word or character embeddings. Although the bidirectional mechanism takes longer to function, it can be negligible since it slightly differs.

**Table 6** Word-level baseline.

| Model | Accuracy | Precision | Recall | Train time | Test time |
|---|---|---|---|---|---|
| Vanilla RNN + (fixed random weight) | 81.788 | 83.140 | 79.323 | **750 s** | **1.59 $\mu s$** |
| Vanilla RNN + (trainable random weights) | 93.715 | 98.542 | 92.418 | 2,650 s | 2.02 $\mu s$ |
| Vanilla RNN + fastText | 93.784 | 98.726 | 92.553 | 3,005 s | 2.46 $\mu s$ |
| LSTM + fastText | 92.689 | **99.059** | 90.619 | 6,814 s | 3.17 $\mu s$ |
| GRU + fastText | 93.869 | 98.880 | **92.572** | 4,403 s | 3.04 $\mu s$ |
| **Bidirectional LSTM + fastText** | **93.898** | 98.887 | 91.720 | 12,090 s | 6.08 $\mu s$ |

**Table 7** Character-level baseline.

| Model | Accuracy | Precision | Recall | Train time | Test time |
|---|---|---|---|---|---|
| Vanilla RNN + (fixed random weight) | 82.467 | 88.679 | 76.595 | **561 s** | **1.39 $\mu s$** |
| Vanilla RNN + (trainable random weights) | 83.918 | 89.100 | 78.400 | 661 s | 1.43 $\mu s$ |
| LSTM + (trainable random weights) | 85.358 | 89.700 | 82.000 | 1,876 s | 1.82 $\mu s$ |
| GRU + (trainable random weights) | 85.438 | 90.056 | 82.469 | 1,805 s | 1.73 $\mu s$ |
| **Bidirectional LSTM + (trainable random weights)** | **86.194** | **90.383** | **83.557** | 3,157 s | 1.92 $\mu s$ |

In this stage, we develop our final architecture by exploiting the effective components of previously-examined methods. Since their accumulation is speed-efficient in our case, both word and character embeddings are utilized simultaneously to gain insight from two different representations and their previously discussed benefits. The two embeddings are further connected to a deep architecture, comprised of LSTM layers that process text bidirectionally before connecting to a TimeDistributed output layer for the token classification purpose. In contrast to unidirectional layers, bidirectional LSTM layers account for textual information from the past and future of a time frame through both forward and backward processing. The effectiveness of this architecture is established by drawing analogies with modern methodologies.

Over the years, several publicly-available Persian spell-checkers have been developed in the industry, but the same is not true in academia. Previous studies' implementations were either not released or not available at the time of writing this paper. This hindered comparisons with previous work but motivated us to establish a baseline for future studies. However, this came at the expense of competing with industrialized spell-checkers, which a plethora of technicians developed using significantly large data and computational resources. Nonetheless, we compared our final approach with four applications. Previous studies have compared themselves with Microsoft Word. Besides, Lilak[9] is another commonly used spell-checker among Persian writers which proofreads text after installing an extension in Mozilla Firefox or Google Chrome. Furthermore, we also compared our approach with two highly-developed applications called Virastman and Paknevis[10]. One dilemma when drawing comparisons is the non-identical set of error classes that these applications identify, including ours. This leaves comparisons to be made on binary classes based on whether a misspelt word is detected or not. We designated a test case consisting of 202 synthetically misspelt and 2,651 correct words (2,853 words in total) extracted from Wikipedia and news (namely, Wikipedia test case). After testing our method and prior work on our test case, we were able to achieve competitive accuracy, precision, and recall scores, while our approach outperformed others on test time (Table 8). Meaning the knowledge of sentences, words, and characters that our architecture has acquired to make a token classification based on our dataset is an appropriate infrastructure for our task of interest. We also achieved competitive results when we tested our model on PerspellData and Shargh, two publicly available spell-checking test cases provided by Dadmatech[11]. Our approach scored higher precision (93.1%) and recall (94.7%) compared to Paknevis and Virastman on the PerspellData test case (other scores were not being reported). It is noteworthy that such results have been obtained using relatively much fewer training data. Although Algorithm 1 is adaptable so that it can be used to expand our dataset up to a much larger size so that results would consequently rise, the lack of computational power impeded us from following up the upswing trend of scores, which was witnessed by training our approach on datasets ranging from 500,000 to 3,450,918.

[9] https://github.com/b00f/lilak
[10] https://paknevis.ir

**Table 8** A comparison between our work and spell-checking industry leaders.

| Model | Accuracy | Precision | Recall | Test time |
|---|---|---|---|---|
| **Wikipedia test case** | | | | |
| Microsoft Word | 89.87 | 90.49 | 98.48 | 1.8 s |
| Firefox Lilak | 96.78 | 97.66 | 98.85 | 2.3 s |
| Virastman | 95.41 | 98.98 | 96.19 | - |
| Paknevis | 98.32 | 99.40 | 98.80 | 170 s |
| **Our work** | **97.62** | **98.83** | **98.61** | **1.6 s** |
| **PerSpellData test case** | | | | |
| Paknevis | - | 83.9 | 94.5 | - |
| Virastman | - | 79.5 | 98.9 | - |
| **Our work** | **-** | **93.1** | **94.7** | **-** |
| **Shargh test case** | | | | |
| Paknevis | - | 78.5 | 75.6 | - |
| Virastman | - | 100 | 60.3 | - |
| **Our work** | **-** | **89.1** | **88.6** | **-** |

## 6. Discussion

The initial inspiration for this research stemmed from the inefficiencies of industrialized spell-checking applications. These applications, still have difficulties detecting certain typographical errors among Persian words promptly. For instance, in OOV cases, Paknevis fails to detect سیالت /sayyaalat/ (fluid), which is correctly written as سیالات. Accordingly, we first separated error detection from error correction and examined means of improving the first in this study. This is still capable of further investigation and is yet to be more efficiently solved. We developed the appropriate dataset, introduced a highly-efficient novel method of solving error detection through the introduced Deep Sequential Neural Network architecture, and processed data at both word- and character-level.

Since no firm baseline was available at the time of conducting this research, industrial applications had to be tested to compare our method's performance with prior works. In comparison, previous approaches either took relatively long to process or were unable to achieve efficient results. We developed an approach that performed faster at inference and maintained competitive scores in performance metrics, given the limited resources we had to perform this academic study. Thus, separating error detection and error correction in order to place more emphasis on the first, has proven beneficial. This way, a greater focus can be placed on the detection phase to develop highly-generalized systems capable of detecting a broader range of Persian misspellings.

[11] https://github.com/Dadmatech/Persian-spell-checkers-comparison

By gathering a rational and diverse set of possible typographical errors, we can better model authors' linguistic and cognitive mistakes rather than unnecessarily complicate the process of spell-checking. Moreover, such methods are hard to debug since a lack of separation between detection and correction makes it difficult to pinpoint where an error may have occurred, for example, when the system suggests the wrong correction. Instead, our approach separates between the two, allowing for further system improvement. Advances in error detection can potentially result in more efficient systems, where the overall success is highly dependent on how well the system executes subtasks at each hierarchy. Our approach to the problem differs from that of the literature in various manners, which will be discussed more thoroughly hereafter.

- **Switching from hard-coding to neural modeling:** The first point of difference hinges on the functionality of our final model. While most Persian spell-checkers iterate through each possible error in a predefined set at each time step, we trained a neural network on typographical errors, so they can be detected with a forward propagation of the network at inference. Additionally, since this model is fed with sentence-level text, it learns how the structure of Persian sentences is formed. Even when a word is misspelt arbitrarily to a large degree, the model can identify what corrections need to be applied to turn the word into a correct one, resulting in a more rational sentence. As an example, the fifth row of the sentence with ID eight in Table 3 shows how the word 'را' is misspelt as 'ژع' using the algorithm. Although the misspelt version has no common letters with its original format, the model has been able to find it. It is shape similarity and substitution errors that need to be taken care of in order to change 'ژع' to 'را'. It fits the sentence more accurately.
- **FarsTypo, the right infrastructure:** Another distinguishing feature of our work is the array of error types that our proposed dataset and methodology can detect together. Our algorithm is designed to detect several other Persian typographical errors besides just insertions, deletions, substitutions, and transpositions which most prior studies have only focused on. Furthermore, the randomness of the algorithm allows the arbitrarily occurrence of various error types in a word, which aids the neural network in generalizing more effectively. Therfore, a broader span of typographical errors is possible to be detected. A further consequence of this algorithm is that misspellings are word-independent, and rule-based methods are likely to fail when applied to the OOV words. While the utilization of both word and character embeddings enables our final model to support the detection of this large volume of possible errors.
- **Deep sequential modeling, the faster alternative:** Among previously proposed methods, the majority of processing is done in the offline phase and during inference, they jointly detect and correct errors using predefined patterns. In contrast, our work models Persian typographical errors on a more general scale. It is important to note that even though these methods are expected to be faster, they still suffer from significant speed drawbacks. As illustrated in Table 8, our approach outperformed all available applications in terms of testing time.

Since Table 8 presents a marginally different set of results, we provide readers with certain scenarios to gain more intuition into how different applications perform. While Persian is a natural language that is fulfilled with combinational nouns such as چرخ و فلک/charkh'o falak/ (Carousel) or سیر و سنجد/sir'o senjed/ (garlic and senjed), industrial applications such as Paknevis failed to detect typographical errors among them. Additionally, there are famous English words that people tend to use by pronouncing them using Persian characters (e.g., writing 'huggingface' as 'هاگینگ فیس'). These words are mostly OOV and systems are thus unable to process them. Moreover, Persian words with English origins such as کیبورد/(Keyboard) and پیتزا /(Pizza) cannot be detected by industrial spell-checkers when mistakenly written as کیبرد and پیزا. In other words, such systems do not detect typographical errors in such misspellings. This facet is significantly different from the previous where the word is written correctly and still counted as OOV. Evidently, the diversity and the generality of our dataset are pivotal factors to let our method's word and character embeddings overcome such scenarios.

## 7. Conclusion

In recent years, there have been significant improvements in Persian spell-checking systems, but most of these advancements are industrial and lack an academic baseline. The present study focuses detecting typographical errors among words to further correct them, which is an important step taken by such systems when processing text. We introduced a large parallel dataset of such errors and a Deep LSTM Network to perform token classification across 51 classes.

We have taken four Persian datasets and added different types of errors to them, resulting in the FarsTypo dataset. FarsTypo is one of the largest typographical error datasets in the Persian language, with more than 3 million words, making it a valuable resource for Persian spell-checking. An error detection model based on Bidirectional LSTM networks is proposed that is capable of detecting word-level errors with an accuracy of 93.89%, a precision of 98.88%, and a recall of 91.72%. Additionally, the model is capable of identifying character-level errors with an accuracy of 86.19%, a precision of 90.38%, and a recall of 83.55%.

Aside from setting a baseline for the task, the results demonstrate how our developed architecture is competitively accurate for error detection and outperforms industrial applications in terms of speed. Future studies will likely use our proposed POS-tagged dataset to develop more efficient spell-checkers by building upon stronger and faster error detectors.

As stated at the beginning of the article, spell-checkers typically involve four successive subtasks, including tokenization, error detection, error correction, and ranking candidates. The ability to perform good corrections depends on the quality of detection. Since error detection is a critical component of the spell checking system, here we have only concentrated on that aspect. Despite the impressive performance of the system developed in this study, no suggestion is provided as to how to correct spelling errors. However, our system can be upgraded to include the correction of mistakes.

**Acknowledgements**
Not applicable.

**Author contributions**
M.D. wrote the main manuscript text and did simulation, modeling and coding. H.F. supervised the work. All authors discussed the results, commented on the manuscript, reviewed drafts of the article, and approved the final draft.

**Funding**
No Funding.

**Availability of data and materials**
The dataset will be published publicly after the publication of the article.

**Declarations**
**Ethics approval and consent to participate**
Not applicable.

**Consent for publication**
Not applicable.

**Competing interests**
The authors declare no competing interests.